\documentclass[11pt]{article}

\usepackage{caption}
\usepackage{subcaption}
\usepackage{paralist}
\RequirePackage{graphics}
\RequirePackage[usenames,dvipsnames]{color}
\RequirePackage[pdflatex]{graphicx}
\RequirePackage[colorlinks,bookmarksopen,pdfpagemode=UseOutlines,
linkcolor=blue, citecolor=blue,urlcolor=blue]{hyperref}
\usepackage[commentsnumbered]{algorithm2e}
\usepackage{algorithmic}
\usepackage{amsmath}
\usepackage{amsfonts}
\usepackage{theorem}
\usepackage{graphicx}
\usepackage{psfrag}
\usepackage{multirow}
\usepackage{tikz}
\usetikzlibrary{shapes}
\usepackage{listings}
\usepackage{wrapfig}

\definecolor{light-blue}{rgb}{0.4,0,0.9}
\newcommand{\starl}[1]{\textcolor{light-blue}{#1}}
\newcommand{\starlkw}[1]{\bf \textcolor{light-blue}{#1}}

\def\A{{\cal A}} 
\def\D{{\cal D}} 
\def\E{{\cal E}} 
\def\I{{\cal I}} 
\def\T{{\cal T}} 
\def\U{{\cal U}} 

\renewcommand{\epsilon}{\varepsilon}

\newcommand {\R}{\mathbb R}
\newcommand {\N}{\mathbb N}

\newcommand{\salg}[1]{\relax\ifmmode {\mathcal F}_{#1}\else ${\mathcal F}_{#1}$\fi}
\newcommand{\msp}[1]{\relax\ifmmode (#1, \salg{#1}) \else $(#1, \salg{#1})$\fi}
\newcommand{\dist}[1]{\relax\ifmmode {\mathcal Dist}\msp{#1}
	\else ${\mathcal Dist}\msp{#1}$\fi}
\newcommand{\subdist}[1]{\relax\ifmmode {\mathcal Subdist}\msp{#1}
	\else ${\mathcal Subdist}\msp{#1}$\fi}

\newcommand{\auto}[1]{{\operatorname{\mathsf{#1}}}}

\newcommand{\Trajeq}{\relax\ifmmode {\mathcal R}_\T \else ${\mathcal R}_\T$\fi}
\newcommand{\Acteq}{\relax\ifmmode {\mathcal R}_A \else ${\mathcal R}_A$\fi}
\newcommand{\noop}{\relax\ifmmode \lambda \else $\lambda$\fi}
\newcommand{\close}[1]{\relax\ifmmode \overline{#1} \else $\overline{#1}$\fi}

\renewcommand{\emptyset}{\O}

\newcommand{\num}[1]{\relax\ifmmode \mathbb #1\else $\mathbb #1$\fi}
\newcommand{\nnnum}[1]{\relax\ifmmode
	{\mathbb #1}_{\geq 0} \else ${\mathbb #1}_{\geq 0}$
	\fi}
\newcommand{\npnum}[1]{\relax\ifmmode
	{\mathbb #1}_{\leq 0} \else ${\mathbb #1}_{\leq 0}$
	\fi}
\newcommand{\pnum}[1]{\relax\ifmmode
	{\mathbb #1}_{> 0} \else ${\mathbb #1}_{> 0}$
	\fi}
\newcommand{\nnum}[1]{\relax\ifmmode
	{\mathbb #1}_{< 0} \else ${\mathbb #1}_{< 0}$
	\fi}
\newcommand{\plnum}[1]{\relax\ifmmode
	{\mathbb #1}_{+} \else ${\mathbb #1}_{+}$
	\fi}
\newcommand{\nenum}[1]{\relax\ifmmode
	{\mathbb #1}_{-} \else ${\mathbb #1}_{-}$
	\fi}

\newcommand{\reals}{{\num R}}                    

\newcommand{\exec}[1]{\relax\ifmmode {\sf Execs}_{#1} \else ${\sf Exec}_{#1}$\fi}
\newcommand{\execf}[1]{\relax\ifmmode {\sf Execs}^*_{#1} \else ${\sf Exec}^*_{#1}$\fi}
\newcommand{\execi}[1]{\relax\ifmmode {\sf Execs}^\omega_{#1} \else ${\sf Exec}^\omega_{#1}$\fi}

\newcommand{\ctrace}[1]{\relax\ifmmode {\sf Ctraces}_{#1} \else ${\sf Ctraces}_{#1}$\fi}

\newcommand{\trace}[1]{\relax\ifmmode {\sf Traces}_{#1} \else ${\sf Traces}_{#1}$\fi}
\newcommand{\tracef}[1]{\relax\ifmmode {\sf Traces}^*_{#1} \else ${\sf Traces}^*_{#1}$\fi}
\newcommand{\tracei}[1]{\relax\ifmmode {\sf Traces}^\omega_{#1} \else ${\sf Traces}^\omega_{#1}$\fi}

\newcommand{\frag}[1]{\relax\ifmmode {\sf Frags}_{#1} \else ${\sf Frags}_{#1}$\fi}
\newcommand{\fragf}[1]{\relax\ifmmode {\sf Frags}^*_{#1} \else ${\sf Frags}^*_{#1}$\fi}
\newcommand{\fragi}[1]{\relax\ifmmode {\sf Frags}^\omega_{#1} \else ${\sf Frags}^\omega_{#1}$\fi}

\newcommand{\reach}[1]{\relax\ifmmode {\sf Reach}_{#1} \else ${\sf Reach}_{#1}$\fi}

\newcommand{\fstate}{{\sf fstate}}
\newcommand{\lstate}{{\sf lstate}}

\newcommand{\sayan}[1]{{#1}}

\definecolor{sayancolor}{rgb}{0.776,0.22,0.07}

	\lstdefinelanguage{xyz}{
		keywordstyle=\bf, 
		identifierstyle=\it, 
		emphstyle=\starl, 
		mathescape=true,
		tabsize=20,
		sensitive=false,
		columns=fullflexible,
		keepspaces=false,
		flexiblecolumns=true,
		basewidth=0.05em,
		moredelim=[il][\rm]{//},
		moredelim=[is][\sf ]{!}{!},
		moredelim=[is][\bf ]{*}{*},
		keywords={
			else,elseif,end,eff,
			for, foreach,forget,
			input,internal,if,imports,in,
			or,
			pre,
			return,
			then,type,types,thread,to,tasks,
			variables, vocabulary,
			when,where, with,while},
		emph={ doMove, doReachAvoid, doReachA, forget, getOdometer, getPos, getxPos, getPosition, getNeighbors, inMotion, sharedsw, sharedmw, tuple, map, array, enumeration},
		literate=
		{(}{{$($}}1
		{)}{{$)$}}1
		{\\in}{{$\in\ $}}1
		{\\preceq}{{$\preceq\ $}}1
		{\\subset}{{$\subset\ $}}1
		{\\subseteq}{{$\subseteq\ $}}1
		{\\supset}{{$\supset\ $}}1
		{\\supseteq}{{$\supseteq\ $}}1
		{\\forall}{{$\forall$}}1
		{\\le}{{$\le\ $}}1
		{\\ge}{{$\ge\ $}}1
		{\\gets}{{$\gets\ $}}1
		{\\cup}{{$\cup\ $}}1
		{\\cap}{{$\cap\ $}}1
		{\\langle}{{$\langle$}}1
		{\\rangle}{{$\rangle$}}1
		{\\exists}{{$\exists\ $}}1
		{\\bot}{{$\bot$}}1
		{\\rip}{{$\rip$}}1
		{\\emptyset}{{$\emptyset$}}1
		{\\notin}{{$\notin\ $}}1
		{\\not\\exists}{{$\not\exists\ $}}1
		{\\ne}{{$\ne\ $}}1
		{\\to}{{$\to\ $}}1
		{\\implies}{{$\implies\ $}}1
		{<}{{$<\ $}}1
		{>}{{$>\ $}}1
		{=}{{$=\ $}}1
		{~}{{$\neg\ $}}1
		{|}{{$\mid$}}1
		{'}{{$^\prime$}}1
		{\\A}{{$\forall\ $}}1
		{\\E}{{$\exists\ $}}1
		{\\/}{{$\vee\,$}}1
		{\\vee}{{$\vee\,$}}1
		{/\\}{{$\wedge\,$}}1
		{\\wedge}{{$\wedge\,$}}1
		{=>}{{$\Rightarrow\ $}}1
		{->}{{$\rightarrow\ $}}1
		{<=}{{$\ \leq\ $}}1
		{<-}{{$\leftarrow\ $}}1
		{==}{{$=\mathrel{\mkern-3mu}=\ $}}1
		{~=}{{$\neq\ $}}1
		{\\U}{{$\cup\ $}}1
		{\\I}{{$\cap\ $}}1
		{|-}{{$\vdash\ $}}1
		{-|}{{$\dashv\ $}}1
		{<<}{{$\ll\ $}}2
		{>>}{{$\gg\ $}}2
		{||}{{$\|$}}1
		{[}{{$[$}}1
		{]}{{$\,]$}}1
		{]]]}{{$]\rangle$}}1
		{<=>}{{$\Leftrightarrow\ $}}2
		{<->}{{$\leftrightarrow\ $}}2
		{(+)}{{$\oplus\ $}}1
		{(-)}{{$\ominus\ $}}1
		{_i}{{$_{i}$}}1
		{_j}{{$_{j}$}}1
		{_{i,j}}{{$_{i,j}$}}3
		{_{j,i}}{{$_{j,i}$}}3
		{_0}{{$_0$}}1
		{_1}{{$_1$}}1
		{_2}{{$_2$}}1
		{_n}{{$_n$}}1
		{_p}{{$_p$}}1
		{_k}{{$_n$}}1
		{@}{{}}0
		{\\delta}{{$\delta$}}1
		{\\R}{{$\R$}}1
		{\\Rplus}{{$\Rplus$}}1
		{\\N}{{$\N$}}1
		{\\times}{{$\times\ $}}1
		{\\tau}{{$\tau$}}1
		{\\alpha}{{$\alpha$}}1
		{\\beta}{{$\beta$}}1
		{\\gamma}{{$\gamma$}}1
		{\\ell}{{$\ell\ $}}1
		{--}{{$-\ $}}1
		{\\TT}{{\hspace{1.5em}}}3
	}
	
	\lstdefinelanguage{xyzNums}[]{xyz}
	{
		numbers=left,
		numbersep=15pt,
		numberstyle=\tiny,
		stepnumber=2,
		numbersep=4pt,
		xleftmargin=2em,
		frame=single,
		framexrightmargin=-1.5em,
		framexleftmargin=1.5em
	}
	
	\lstdefinelanguage{xyzNumsRight}[]{xyz}
	{
		numbers=right,
		numbersep=15pt,
		numberstyle=\tiny,
		stepnumber=2,
		numbersep=4pt,
		xleftmargin=2em,
		frame=single,
		framexrightmargin=-1.5em,
		framexleftmargin=1.5em
	}

\title{\LARGE \bf Porting Code Across Simple Mobile Robots\thanks{.}}

\author{Yixiao Lin, Sayan Mitra and Shuting Li \\
	Coordinated Science Laboratory \\
University of Illinois at Urbana-Champaign}

\begin{document}
\maketitle

\begin{abstract}
The StarL programming framework aims to simplify development of distributed robotic applications by providing easy-to-use language constructs for communication and control.  It has been used to develop applications such as  formation control, distributed tracking, and collaborative search. In this paper, we present a complete redesign of the StarL language and its runtime system which enables us to achieve portability of robot programs across platforms. Thus, the same application program, say, for distributed tracking,  can now be compiled  and deployed on multiple, heterogeneous robotic platforms.
Towards portability, this we first define the semantics of StarL programs in a way that is largely platform independent, except for a few key platform-dependent parameters that capture the worst-case execution and sensing delays and resolution of sensors. 
Next, we present a design of the StarL runtime system, including a robot controller, that meets the above semantics. The controller consists of a  platform-independent path planner implemented using RRTs and a platform-dependent way-point tracker that is implemented using the control commands available for the platform.
We demonstrate portability of StarL applications using  simulation results for two different robotic platforms, and several applications.

\end{abstract}

\paragraph*{keywords}
Mobile robots; 
Programming;
Semantics; 
Runtime system
%
	

\section{Introduction}
\label{sec:intro}

Distributed robotic systems have been  at the frontier of research in  manufacturing~\cite{dilts1991evolution,Correll08}, transportation~\cite{naturenews_platoon2015}, logistics~\cite{kivaFORBES,WurmanDM08}, and exploration~\cite{burke2004moonlight}. Yet, managing and experimenting with distributed robotic platforms remains a daunting task. Deploying a new algorithm on a  hardware platform  typically  takes weeks. It  takes months to port an existing application program to a new platform with different dynamics or to deploy it on a heterogeneous mix of vehicles. Compare these efforts with the effort needed to perform comparable tasks for desktop applications. This high overhead cost takes a toll on the robustness of the scientific claims and repeatability of experiments. 

We are developing the Stabilizing Robotics Language (StarL)~\cite{DBLP:conf/lctrts/LinM15} and its supporting APIs and runtime system to mitigate some of these issues.  StarL provides programming language constructs for coordination and control across robots. See for example, the StarL program for in Section~\ref{sec:flocking} for making a collection of robots form a regular polygon. 
The design goal for this language is to make robot programming  faster, more abstract,  cleaner, and closer to high-level   textbook pseudo-code~\cite{Magnusbook2010,Cortes02}. The result would be programs that are also easier to debug and verify. 

Two key features of StarL are a distributed shared memory (DSM) primitive for coordination and a reach-avoid primitive for control. DSM allows a program to declare  program variables that are shared across  multiple robots (e.g. $\starlkw{sharedmw}$ $x: Int$). This enables programs running on different robots to communicate by writing-to and reading from the shared variable $x$. The runtime system implementing the DSM ensures that the updated values of $x$ are consistently propagated to appropriate robots using message passing. The reach-avoid primitive provides a function $\starl{doReachAvoid(x,U)}$ and a set of control flags. A StarL program calling $\starl{doReachAvoid(x,U)}$ instructs the low-level robot motion controller to reach the target point $x$ while avoiding the region $U$ (defined in a common coordinate system). The control flags are updated by the low-level controller to indicate status information such as the target has been reached, the region $U$ has been violated, or that the controller is giving-up.

In this paper, we present  a newly implemented feature of StarL, namely, portability across different  platforms. For some platforms (e.g., an industrial robotic arm vs. a bipedal robot),  there is little commonality in the task and their implementations, and therefore, portability is not meaningful. However,  there exists a class of distributed robotic platforms and related tasks, such as, visiting a sequence of points in 3D space to achieve some higher-level goal, where automatic portability of can improve both quality of programs and the productivity of developers. 
Consider a distributed collaborative search application. The high-level code deals with leader election, allocation of subtasks to robots, and tracking failures and progress. While individual participating robots may have different low-level controllers for tracking way-points, the higher-level code can be made dynamics-independent, and therefore, portable. 
In this paper, we present the design of the StarL language and supporting runtime system which demonstrate feasibility of this idea. The implementation of our StarL framework is available at \url{https://github.com/lin187/StarL1.5}

\subsection{Overview of contributions}
\label{sec:contrib}
First, we define what it means to port a StarL program across platforms. When two instances $P(m_1)$ and $P(m_2)$ of the same program  $P$ are executed on two different desktop computers $m_1$ and $m_2$ (with different hardware, OS, and execution environments), we expect the {\em observable behaviors they produce to be semantically equivalent\/}, while there may be  other aspects of their behaviors (say, execution time) that differ.
Analogously, in order to discuss portability of robot programs, first we define the platform-invariant semantics of StarL. 
Mathematically, we view the evolution of a StarL program together with its hardware stack and physical environment, as a hybrid automaton. 
Within this framework, we define semantics in terms of the relationship between the calls to $\starl{doReachAvoid}$ and the corresponding changes to the control status flags. A crucial difference in our semantics compared to the semantics of ordinary programs is that, by necessity, our semantics does not completely abstract away timing and position information. 

Our second contribution is the design and implementation of a runtime system that enables porting programs across robotic platforms. A StarL program consists of several threads that interact through a part of memory called the variable holder. When a StarL program calls $\starl{doReachAvoid}(x,U)$, it instructs to the robot controller that the new target position is $x$ and a region to be avoided is $U$. The controller has two continuously executing threads: a path planner and a way-point tracker. The path planner plans possible paths from the current position to the target (avoiding U) and provides a sequence of way-points to the way-point tracker. The path planner may fail to discover a path. 


%
The StarL compiler translates the application program, with additional platform specific information, to executable code for each target platform.
The same high-level code could be compiled for different platforms and instrumented for simulations and monitoring.
We demonstrate this capability with several applications and two popular platforms. Our experimental results use  the a discrete event simulator that runs actual StarL executable programs (Java) with detailed, physics-based simulations of the platform and the communication channels.

\subsection{Related work}
\label{sec:related}
Several projects have pushed the limits of custom-built software and hardware for distributed robotics  (see~\cite{konolige2004centibots,rubenstein2012kilobot,chaimowicz2005controlling} for some recent examples); this approach is orthogonal to our objective of developing portable software. 

There are several libraries for programming robots using  languages like Python and Matlab~\cite{quigley2009ros,gerkey2003player,blank2003python,corke1996robotics,nesnas2007claraty,nesnas2007claraty2,calisi2008openrdk,hotdec1,hotdec2} and notably the Reactive Model-based Programming Language (RMPL)~\cite{williams2003model}, but they do not provide high-level coordination and control APIs  nor do they address portability.
%

The Robotics Operating System (ROS)~\cite{quigley2009ros} provides a library of drivers and functions for programming robots and has an ever-growing community of developers and users. While parts of ROS library is reusable, there is currently no effort towards automatically porting programs across platforms.
Our previous work on StarL~\cite{DBLP:conf/lctrts/LinM15,zimmerman2013starl} provided software components for  mutual exclusion,  leader election, service discover, etc. but also did not address portability or the semantic issues. 
%
%

Our point of view for programming differs from the correct-by-construction synthesis approach (for example,~\cite{DBLP:journals/trob/Kress-GazitFP09,DBLP:conf/icra/DingLHT11,DBLP:journals/corr/SvorenovaKCCCB14,DBLP:journals/tac/GolDLB14,DBLP:journals/tac/WongpiromsarnTM12,DBLP:conf/icra/NedunuriPMCK14,DBLP:conf/popl/BeyeneCPR14}) in what we see as the role of the user. A correct-by-construction synthesis algorithm takes as input a high-level requirement to generate  robot programs for accomplishing this requirement. 
In our approach, human creativity continues to be central in writing the step-by-step program that solves the task, but  the tedious and error-prone steps in low-level coordination and  control are automated, and specifically in this paper, we focus on enabling  automatic porting of programs across platforms.

\section{Preliminaries and Problem Setup}
\label{sec:problem}

\subsection{Example: Formation program}
\label{sec:flocking}
The simple formation program  in Figure~\ref{fig:gen_formation} illustrates some of the key features of the StarL language. 
The program encodes a heuristic which when executed by  an (odd) number of robots form a regular polygon. This StarL program is transformed by our compiler to Java program that is executed at each robot (see,~\cite{DBLP:journals/corr/LinM15} for more details on this transformation). 
The program has two parts: variable declarations and  guarded-commands. 

\paragraph*{variables}
When the program is executed by robot $i$, the scope of the first four variables is only the program at robot $i$. 
The variable $\mathit{pid}$ is initialized with a call to the built-in function $\starl{getId}$ which returns a unique identifier for each participant robots.
The variable $\mathit{loc}$ is used like a program counter.
$\mathit{ItemPosition}$ is a built-in type for storing 3D coordinates of points in space with respect to a common and fixed coordinate system.

The $\starlkw{sharedsw}$ keyword declares $\mathit{pos}[i]$ as a single-writer (sw) multi-reader shared variable. That is, the scope of the integer array  $\mathit{pos}[]$ spans all  instances of formation program executing, all its components can be  read by all instances, but only the instance running at robot $i$ can write to $\mathit{pos}[i]$.
The StarL runtime system implements a distributed share memory (DSM) system over wireless message passing. For example,   whenever the program at $i$ writes a new value, say $5$, to a shared variable $x$ with the assignment $\mathit{x = 5}$, the runtime system detects this change, then sends $\langle x, 5, \mathit{timestamp}\rangle$ to other participants running the same application, and upon receiving these messages, the runtime system at the other end updates its copy of $x$ to $5$.

\paragraph*{guarded-commands}
The main part of the program consists of several guarded-command ($\starlkw{pre-eff}$) blocks. These blocks are translated to {\bf if} conditionals inside  a big {\bf while} loop in the transformed Java program. In each iteration of this big loop, the $\starlkw{eff}$ect part of a block may be executed if its $\starlkw{pre}$condition holds. If multiple preconditions hold (and no priority order is specified) then one of the effects is executed arbitrarily.

\begin{figure}
	\begin{center}
		\scriptsize
\begin{lstlisting}[language=xyzNums]
pid: Int = $\starl{getId}$();  $\label{ln:initG}$ 
loc: enum{init, calc, wait} = init;
counter: Int = 0;
target: ItemPosition;
sharedsw pos[pid]: ItemPosition; 
 
initialize()
	pre loc == init
	eff pos[pid] =  $\starl{getPos}()$; loc = calc;

update()
	pre loc == calc ||  $!\starl{active}$
	eff target = bisector(pos,pid,len);
		$\starl{doReachAvoid(target,0)}$;
		counter = 0; loc = wait;

wait()
	pre loc == wait $\&\&$ counter $<=$ 5
	eff pos[pid] =  $\starl{getPos}()$;
		counter++;
		if counter > 5 then loc = calc 
			
\end{lstlisting}
		\caption{StarL code of the formation program.}
		\label{fig:gen_formation}
	\end{center}
\end{figure}

The formation program has three blocks. The $\mathit{initialize}()$ block is executed once at the beginning and it sets $\mathit{pos}[i]$ to be the current position of the robot using built-in $\starl{getPos}$ function.
It also sets $\mathit{loc = calc}$ which ensures that only the $\mathit{update}()$ block can execute next. 
The next block $\mathit{update}()$ is enabled when $\mathit{loc} = calc$ or the special boolean variable $\starl{active}$ is set to ${\sf false}$. It computes a target position for the robot using the $\mathit{bisector}$ function (not shown here). This function when executed by robot $i$ computes a point (at distance $\mathit{len}$) on the perpendicular bisector of the line joining $\mathit{pos}[j]$ and $\mathit{pos}[(j+1) \% n]$, where $j$ and $(j+1) \% n$ are the robots diametrically opposite to $i$.
It is straightforward to check that if $\mathit{pos}[]$ array forms a regular $n$-sided polygon then the program reaches an equilibrium. The $\starl{doReachAvoid(target,0)}$ function call communicates to the robot's controller that the new destination is $\mathit{target}$. 
Generally, $\starl{doReachAvoid(x,U)}$ takes two arguments:  $x$ is a point in space and $U$ is a set. Roughly, this call communicates to the robot controller to reach $x$ while avoiding $U$. In the remainder of the paper, we will discuss the precise semantics of $\starl{doReachAvoid}$, its interaction with StarL applications, and its portable implementations.  
Finally, the $\mathit{wait}()$ block  merely updates the $\mathit{pos}[i]$ variable and the $\mathit{counter}$ so that the 
$\mathit{wait}$ block is executed 5 times after every execution of $\mathit{update}$. 

\subsection{Semantics of StarL applications}
\label{sec:reach-avoid-semantics}

When a StarL program (such as the one in Figure~\ref{fig:gen_formation}) is ported from one platform to another, we expect that certain important aspects of the behavior of the program to be preserved, while other less crucial aspects may alter. 
In this section, we discuss the semantics of StarL programs in terms of the expected behavior of certain function calls and variables. This will be the basis for specifying what it means to correctly port StarL programs across different robot platforms.

Semantics of a StarL applications is defined using hybrid I/O automata  (HIOA)~\cite{Mitra07PhD,TIOAmon}. HIOA  is a mathematical framework for precisely describing systems that have both discrete and continuous dynamics. Discrete dynamics is modeled by {\em transitions\/} that instantaneously change the state variables of the system. Typically transitions are defined by program statements that change state variables associated with the computing stack. Continuous evolution is modeled by {\em trajectories\/} that govern how state variables evolve over an interval of time. A trajectory is a function that maps an interval of time to valuations of variables. Typically they are specified using differential and algebraic equations and define the evolution of the physical variables (variables related to the computing stack remain constant over a trajectory). 

All the program threads implementing the application, the message channels, as well as the physical environment of the application (robot chassis, obstacles) are modeled as hybrid automata, and the overall system is described by a giant composition of these automata (see Figure~\ref{fig:robot-gvh-env}). The semantics of the control API is then defined in terms of the allowed observable behaviors or traces of $\auto{system}$.
A complete formal presentation of the semantics is beyond the scope of this  paper, we refer the interested reader to the publications on virtual nodes which show another application of  HIOA framework in giving precise semantics of a distributed programming system~\cite{GLMN:TAAS09}.

A {\em hybrid I/O automaton\/} is a tuple $\A = \langle V, \Theta, A, \D, \T \rangle$,
where 
\begin{inparaenum}[(i)]
\item $V$ is a set of variables partitioned into input, output and state (internal) variables; set of all possible valuations of the state variables $Q$ is the set of states. 
\item $\Theta \subseteq Q$ is a set of initial states, 
\item $A$ is a set of actions partitioned into input, output, and internal actions,
\item $\D \subseteq Q \times A \times Q$ is the set of discrete transitions, and 
\item $\T$ is a well-defined set of trajectories for all the variables $V$.
For any trajectory $\tau \in \T$, $\tau.\fstate$ is the first state and if $\tau$ is closed then $\tau.\lstate$ is the last state of $\tau$.

\end{inparaenum}
In addition, it is required that $\A$ satisfies the input action and input trajectory enabled condition~\cite{Mitra07PhD,TIOAmon} which implies that it cannot block (a) any input action triggered by another automaton in the system (possibly the environment), and (b) it cannot block any input trajectory produced by another automaton in the system. 

A particular behavior of $\A$ is called an {\em execution\/} and mathematically it is  an alternating sequence $\alpha = \tau_0, a_1, \tau_1, a_2, \ldots$, such that $\tau_0.\fstate \in \Theta$ and $(\tau_i.\lstate, a_{i+1}, \tau_{i+1}.\fstate) \in \D$. In general, HIOA are open and nondeterministic, and therefore, have many executions from the same starting state.
Once we fix an execution $\alpha$ of an HIOA, we can use the notation $v(t)$ to denote the valuation of a variable $v \in V$ at time $t$ in that execution $\alpha$.

Each StarL application executes several program threads, and interacts with the environment through different sensors and actuators.
Figure~\ref{fig:robot-gvh-env} shows these different component HIOAs that make-up the model of a StarL application and its environment (the overall HIOA model is obtained by composing the component HIOAs). 
The HIOA models for the individual  threads capture the evolution of the corresponding program variables according to the guarded commands in the program. Time elapses between the successive transitions, but the exact duration of time is unpredictable and platform dependent. The controller uses special program threads (also modeled as an HIOA) responsible for implementing reach-avoid control strategies. The variable holder is a part of the runtime system that maintains shared variables and message buffers.

All the threads communicate with each other and with the sensors and actuators through variables in the variable holder. For example, the $\mathit{currentPos}$ variable is updated periodically by the positioning sensors (e.g. GPS) and read by the application programs by calling the $\starl{getPos}()$ function.
For this paper, we will focus on the control API of StarL which consists of the following single-writer multi-reader shared variables stored in the variable holder:  

\begin{figure}
	\begin{center}
		\includegraphics[scale=0.33]{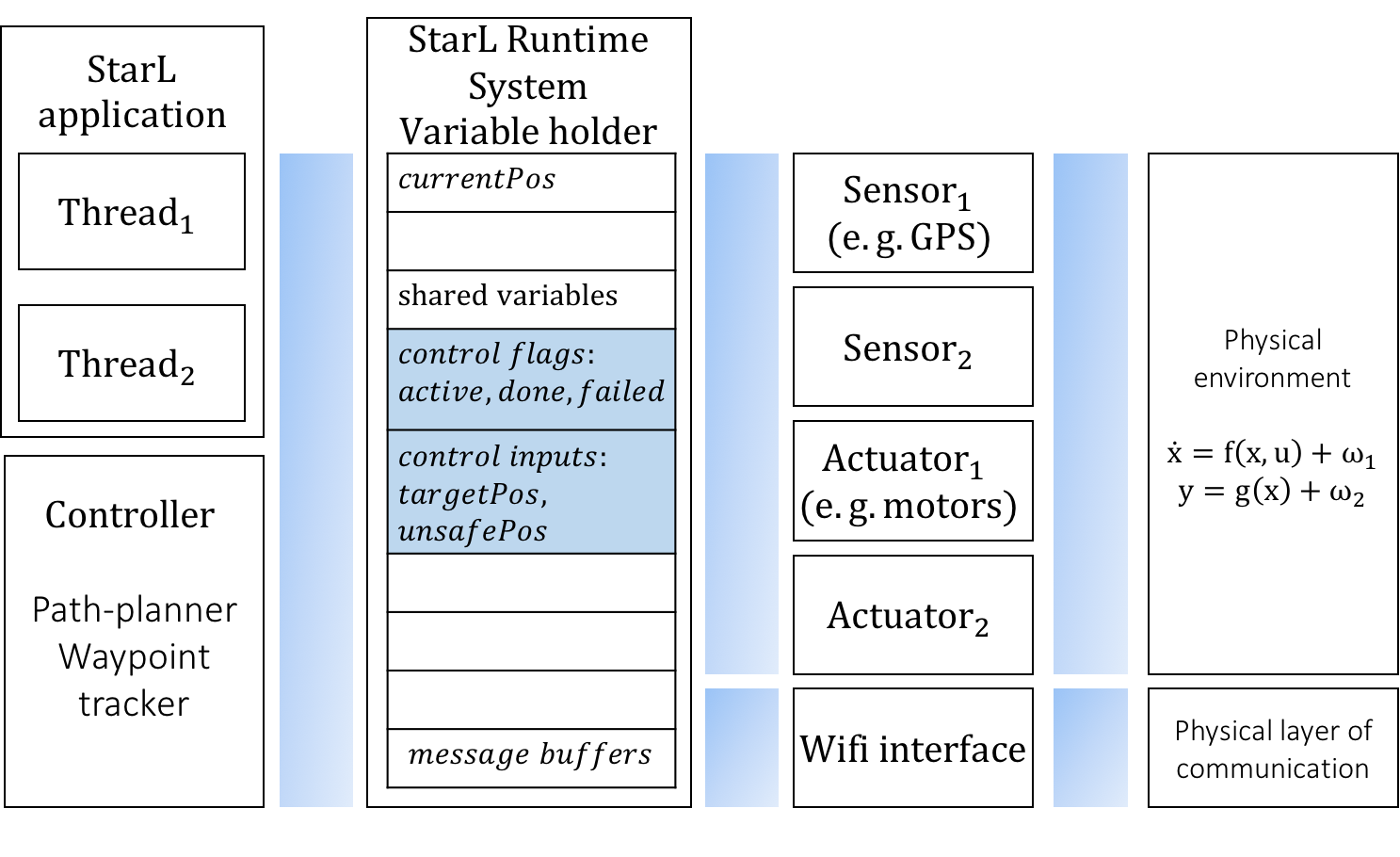}
	\end{center}
	\caption{\scriptsize Application threads, controller, variable holder, and physical environment. Shaded areas indicate information flow through messages, sensors, and actuators.}
	\label{fig:robot-gvh-env}
\end{figure}

\begin{table}[h!]
	\label{controlAPIvars}
	\begin{tabular}{|l|l|l|}
		\hline 
		Shared variable & Writer & Interpretation \\
		\hline 
		$\mathit{targetPos}$  & App. & coordinates of target position \\
		$\mathit{currentPos}$ & Sensor & coordinates of last recorded position \\
		$\mathit{unsafePos}$   & App. & set of coordinates of unsafe positions \\
		$\mathit{active}$ &  Ctrl. & attempting progress towards target\\
		$\mathit{done}$ &  Ctrl. & reached vicinity of target\\
		$\mathit{failed}$ & Ctrl. & cannot make progress or entered unsafe \\		 
		\hline
	\end{tabular}
	\caption{\scriptsize Single-writer multi-reader shared variables in the control API. Each of the variables are  written to by one of the following threads: the main application thread (App.), the controller thread (Crtl.), and sensor update threads.}
\end{table}
Each of these variables are written to by a single thread as indicated in the table. When a call to \starl{doReachAvoid(x,U)} from the application program is executed, it sets:
\begin{lstlisting}[language=xyz]
targetPos := x;
unsafePos := U;
\end{lstlisting} 
thus communicating to the controller thread that the new target is $x$ and the unsafe region is $U$. Once this call is executed, the control thread attempts to move the robot towards $x$ while avoiding $U$ and updates the three status variables $\mathit{active}$, $\mathit{done}$, and $\mathit{failed}$. 

 With this set-up, we can now define the key semantic properties of StarL programs.
 In all of the following statements, we fix any execution $\alpha$ of the  HIOA describing the system,  we assume that at some time $t_0 >0$ a $\starl{doReachAvoid(x,U)}$ call completes execution and there are no other calls to $\starl{doReachAvoid}$ between $t_0$ and another arbitrary time point $T > t_0$. Semantics of the control flags are specified in terms of two platform dependent parameters: a dwell time $d_t>0$ and a quantization distance $q_d>0$.
 We define the predicate $reached(t) := d(\mathit{currentPos}(t_1),\mathit{targetPos}(t)) \leq q_d$, that is, the current recorded position of the robot in the variable holder is within $q_d$ of the last target position issued by invocation of $\starl{doReachAvoid}$.

\scriptsize
\begin{align*}
\mathit{D1:} & \exists \ t \in [t_0, T]:  \mathit{done}(t) \implies \exists \ t_1 \in [t_0,  t]:  \mathit{reach}(t_1). \\ 
\mathit{D2:} & \exists \ t_1 > t_0,  \forall \ t \in [t_1, t_1 + d_t], \mathit{reach}(t) \\
	& \implies \exists t_2 > t_1, \forall \ t' \in [t_2, T], \mathit{done}(t').
\end{align*}

\normalsize 
Note the asymmetry in the two conditions. 
D1 states that along any execution $\mathit{reach}$ precedes the corresponding $\mathit{done}$. 
D2 states that along any execution, if $\mathit{reach}$ {\em persists\/} for at least $d_t$ duration (and no further $\starl{doReachAvoid}$'s are issued) then $\mathit{done}$ becomes set. Persistence is required for the runtime system to detect predicates.
Next, we define the predicate $crossed(t) := d(\mathit{currentPos}(t_1),\mathit{unsafe}(t)) \leq q_d$, that is, the current recorded position of the robot is within $q_d$ distance of an unsafe position issued by invocation of $\starl{doReachAvoid}$.

\scriptsize
\begin{align*}
\mathit{F1:} & \exists \ t \in [t_0, T], \mathit{failed}(t)) \implies \ \exists \ t_1 \in [t_0, t], \mathit{crossed}(t) \\
\mathit{F2:} & \exists \ t_1 > t_0,  \forall \ t \in [t_1, t_1 + d_t], \mathit{crossed}(t) \\
& \implies \exists t_2 > t_1, \forall \ t \in [t_2, T], \mathit{failed}(t).
\end{align*}

\normalsize 

F1 states that along any execution $\mathit{crossed}$ precedes the corresponding $\mathit{failed}$. 
F2 states that along any execution, if $\mathit{crossed}$ {\em persists\/} for at least $d_t$ duration (and no further $\starl{doReachAvoid}$'s are issued) then $\mathit{failed}$ becomes set. 
In addition to the $\mathit{done}$ and $\mathit{failed}$ flags, the controller API provides the $\mathit{active}$ status flag for the low-level controllers to communicate with the application programs.

\scriptsize
\begin{align*}
\mathit{A1:} & \exists \ t \in [t_0, T], \mathit{active}(t)  \implies \forall  \ t_1 \leq t, \neg \mathit{done}(t_1) \wedge  \neg \mathit{failed}(t_1). 
\end{align*}

\normalsize 

A1 implies that at  any point in an execution, if $\mathit{active}$ is set then it is neither preceded by $\mathit{done}$ nor by $\mathit{failed}$. 
The semantics of $\mathit{active}$ being set to false is unspecified. Informally, this means that the low-level controller has given-up on attempting to reach the $\mathit{targetPos}$, but the the precise conditions under which this can happen may depend on the platform, the implementation of the controllers, and the environment which may include moving obstacles. 


\subsection{Semantics of porting}
\label{sec:porting}

When a StarL application is ported correctly from one platform to another, we expect the above conditions to be preserved albeit with different (known) values of the platform dependent constants $d_t$ and $q_d$.  
We have decided to state the semantics in terms of program statements and the control flags. This leaves a necessary gap between these statements and the actual physical state of the robot.
(1) The $\mathit{done}$ and $\mathit{failed}$ flags imply conditions only on the recorded position of the robot and not its physical position. Under additional assumptions about the veracity of sensors and underlying dynamics, we can infer that indeed $\mathit{done}$ implies that the actual position of the robot is $q_d$-close to the target issued.
(2) Conversely, under additional assumptions about persistence of the actual position of the robot near the target and the speed of the runtime system, we can infer that if the real position is close to the target then $\mathit{done}$ will be set. Similar arguments can be constructed for reasoning about the unsafe set and the $\mathit{failed}$ flag. 

In the next section, we describe our design of the StarL  controller thread and  runtime system which enables us to correctly and automatically port StarL applications across  different platforms. 

\section{Implementation}
\label{sec:solution}

A controller for a robot gets inputs from other StarL programs through the variables
$\mathit{targetPos}$ and $\mathit{unsafePos}$,
and produces outputs by writing to the status variables 
$\mathit{done}, \mathit{failed}$ and $\mathit{active}$.
It may receive input data from other variables and sensors as well.
A valid controller should guarantee the semantics given in the previous section. 
In this section, we describe our implementation choices. 
Our implementation involves two separate threads:
(a) a path planner and a 
(b) a way-point tracker.

\subsection{Way-point tracker}
\label{sec:waypoint-tracker}

The way-point tracker reads inputs $\mathit{currentPos}$ and a way-point (generated by the path planner),
and uses low-level control instructions and a dynamical model of the robot,  
to attempt closed-loop tracking of the way-point. 
It is oblivious to obstacles and the unsafe set. 
Like device drivers, this program has to be written for each platform independently as both the low-level control instructions 
and the dynamics of the model are strongly platform dependent. 
For example, the low-level control instructions and the dynamical models for the 
AR Drone2 and the iRobot Create are shown in Tables~\ref{tab:llardrone} and~\ref{tab:mod_ardrone}. The commands for setting YawSpeed, Pitch, Roll, Gaz can be used in any combination while other commands can only be used one at a time.
\begin{table}[h!]
	\centering
	\scriptsize
	\begin{tabular}{|l|l|}
		\hline
		AR Drone2 & iRobot Create  \\
		\hline
		takeOff()  & straight($v_{\text{ref}}$) \\ 
		land() & turn($a_{\text{ref}}$) \\ 
		hover() & curve($v_{\text{ref}},r$) \\ 
		\hline
		setYawSpeed($a_{\text{ref}}$) & \\ 
		setPitch($\theta_{\text{ref}}$)  & \\ 
		setRoll($\phi_{\text{ref}}$)  & \\ 
		setGaz($v_{\text{ref}}$) & \\ 
		\hline
	\end{tabular}
	\caption{\scriptsize Low-level control instructions available for ARDrone2 and the iRobot Create.}	
      \label{tab:llardrone}
\end{table}
For the AR Drone2 model, the state variables are $x, y, z$ position coordinates, yaw ($\psi$), pitch ($\theta$), roll ($\phi$) angles and the corresponding velocities. 
For the iRobot model the state variables are $x, y$ position coordinates and the heading ($\theta$) angle.

\begin{table}[h!]
	\centering
	\scriptsize
	\begin{tabular}{|l|l|}
		\hline
		AR Drone2 & iRobot Create  \\
		\hline
		 $thrust = (gaz+10)/\cos{\phi}/\cos{\theta}$ & $\dot{x} = v\cdot \cos{\theta}$ \\ 
		$\ddot{x} = -(thrust)(\sin{\phi}\cdot\sin{\psi} + \cos{\phi}\cdot\sin{\theta}\cdot\cos{\psi})/M$ &  $\dot{y} = v\cdot \sin{\theta}$ \\ 
		$\ddot{y} = (thrust)(\sin{\phi}\cdot\cos{\psi} + \cos{\phi}\cdot\sin{\theta}\cdot\sin{\psi})/M$ & $\dot{\theta} = a_{\text{ref}}$ \\ 
		$\dot{z} = gaz$ & \\
		$\dot{\psi} = gain\cdot(a_{\text{ref}} - \psi)$  & \\ 
\hline
	\end{tabular}
	\caption{\scriptsize Dynamical models of AR Drone2 and iRobot Create.}
	\label{tab:mod_ardrone}
\end{table}
Since control commands are provided by the robotic platforms as black boxes, in creating the model for these controller we use standard proportional controllers. For example, the yaw speed control is shown in the last row and similar controls for pitch and roll are used (omitted in the table). This technique can be extended to other closed loop dynamics such as PID. The way-point tracker reads sensor inputs and uses the motion commands in a closed-loop. 

\subsection{Safe path planner}
\label{sec:path-planner}

The path planner attempts to generate a sequence of way-points from $\mathit{currentPos}$ leading to $\mathit{targetPos}$ such that if the robot follows this path it will avoid the set of locations specified by $\mathit{unsafePos}$. It uses a platform-independent implementation of RRT algorithm~\cite{6631297}, instantiated with platform dependent way-point controller. 

Whenever a change in the  $\mathit{targetPos}$ or  $\mathit{currentPos}$ is detected, the following algorithm for growing an RRT is executed with an initial tree that has a single node at $\mathit{currentPos}$. Given $\mathit{currentPos}, \mathit{targetPos},$ $\mathit{unsafePos}$ and a current tree $T$ in $\reals^3$, the RRT algorithm adds new points to the tree as follows:
It picks a random point $\vec{x} \in \reals^3$ and attempts to add it to $T$. 
In doing so, it first picks a point in the tree $\vec{p} \in T$ that is closest to  $\vec{x}$.
Then it simulates the way-point tracker with a model of the plant to go from $\vec{p}$ to $\vec{x}$ to generate a simulated path.
Finally, it checks if this simulated path is sufficiently far from $\mathit{unsafePos}$.
If this check succeeds then it adds $\vec{x}$ to the tree.
Otherwise, it finds another point halfway between $\vec{p}$ and $\vec{x}$ unless the distance between the two falls below a threshold in which case a new $\vec{x}$ is picked at random.

In this way, once the tree $T$ reaches a threshold size the RRT construction algorithm is stopped. If a path from the vicinity of $\mathit{currentPos}$ to the vicinity of $\mathit{targetPos}$ exists in $T$ then this path is sent to the way-point tracker. Otherwise, the $\mathit{active}$ flag is set to $\mathit{false}$. 
Assuming that $\mathit{currentPos}$ and $\mathit{targetPos}$ did not change during the tree construction, this indicates to the StarL application program that a safe path has not been discovered by the path planner. In general, establishing nonexistence of safe path is challenging, and our design of the runtime system leaves it to the programmer to code best effort strategies by detecting when all the control flags $\mathit{active}$,  $\mathit{done}$, and $\mathit{failed}$ simultaneously become $\mathit{false}$.

%
%

\begin{figure}
	\begin{center}
		\includegraphics[scale=0.55]{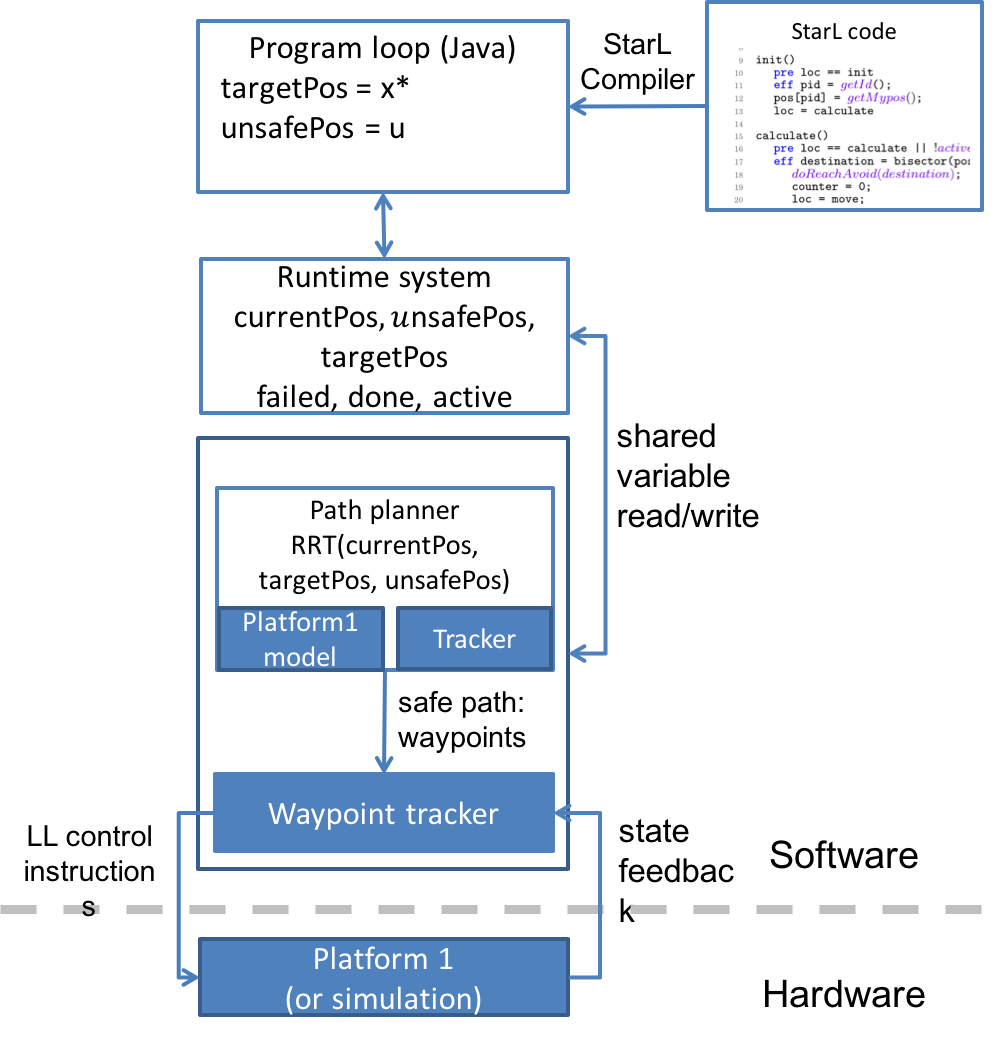}
	\end{center}
	\caption{\scriptsize Porting StarL applications. The shaded blocks are platform dependent and the rest are generic.}
	\label{fig:arch}
\end{figure}




%
%
%
%
%

\section{Simulation experiments}
\label{sec:experiments}

\sayan{StarL applications have been deployed on two different types of robotic platforms at UIUC and UT Arlington: iRobot Create ground robots controlled by  Android smart phones~\cite{DuggiralaJZM12} and AR Drone quadrotors\footnote{https://www.youtube.com/watch?v=27NCnLYHAtY}. StarL also has  a powerful discrete event simulator  that can run the same application code with dozens of robot instances, a simulated physical environment with detailed physics models, and with different kinds of robots. In this section, we report on semantics and portability of StarL applications with simulation experiments.} 

To port a StarL application to a new  platform, developers need to have the appropriate drivers for the hardware (e.g. wifi interfaces, sensors, positioning system, etc.), write the way-point tracker, and provide the physics model of the platform. These are the platform-dependent components shown in blue in Figure~\ref{fig:arch}. 
For our iRobot+Android platform, for example, 
the smartphones handle  the computation and communication (using UDP sockets) and read the sensor data from the iRobot sensors, and write actuation commands to the motion controllers. 
%
%


\subsection{Experiments on semantics of ReachAvoid}
\label{sec:basicReach-avoid}

\begin{figure}
	\begin{center}
		\includegraphics[scale=0.17]{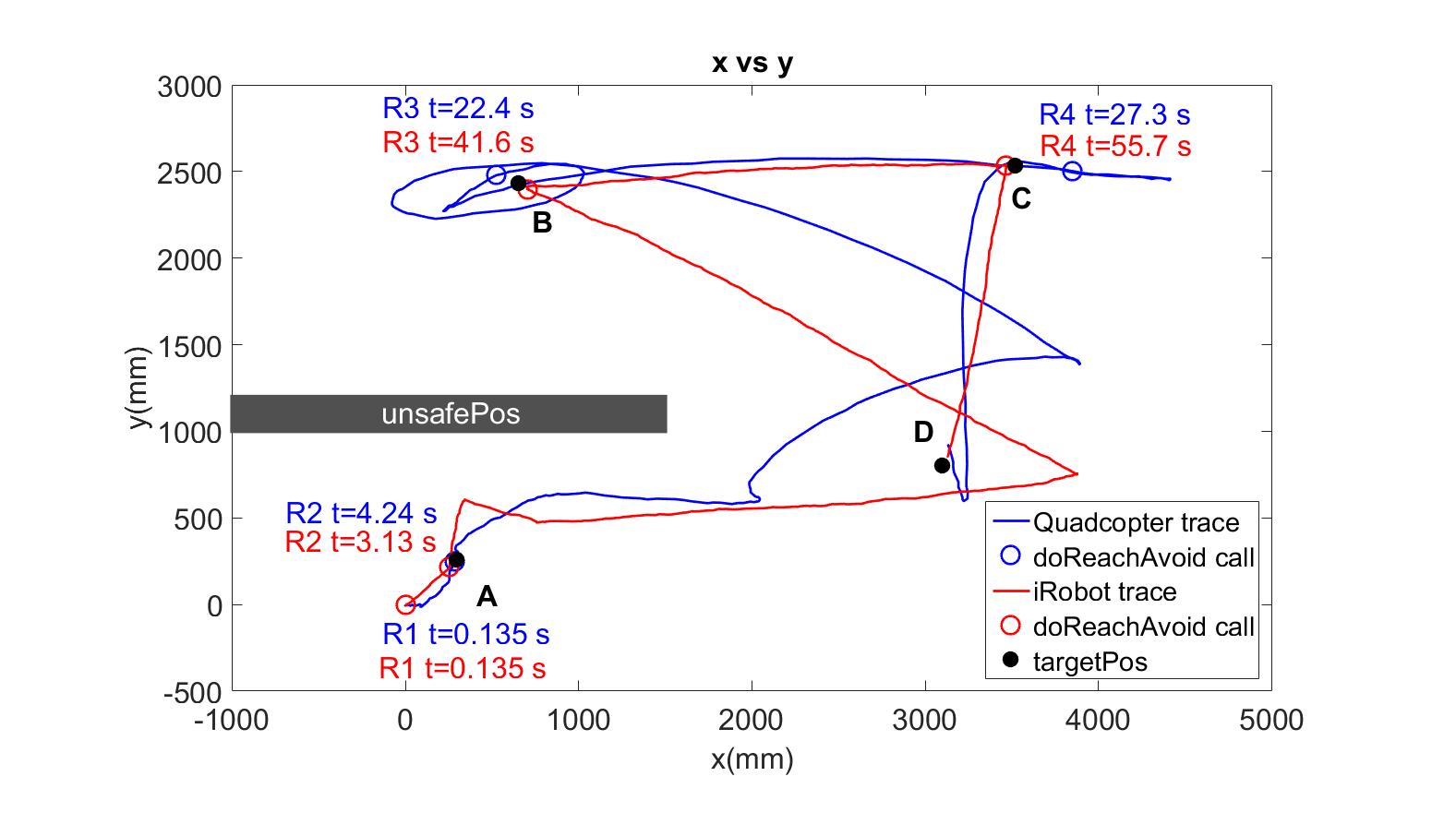}
	\end{center}
	\caption{\scriptsize Two independent  traces of a quadcopter (blue) and a iRobot (red) following a fixed sequence of four way-points (solid circles) with a fixed unsafe set (rectangle). The hollow circles are points where $\starl{doReachAvoid}$ is called. }
	\label{fig:both}
\end{figure}

Our first set of experiments illustrate the semantics of $\starl{doReachAvoid}$ for individual robots in a static environment.
A single robot executes a simple program that is a sequence of four $\starl{doReachAvoid}(x[i], U)$ calls. That is, once the $\starl{done}$ flag is set after the first call, then $\starl{doReachAvoid}(x[i+1], U)$ is called and so on. Here $U$ is a wall as shown in Figure~\ref{fig:both} and we call the four points $A, B, C,$ and $D$.

Two independent traces of running this StarL program on the iRobot (red) and the AR Drone (blue) models are shown in Figure~\ref{fig:both}. 
The hollow circles labeled $Ri t=xx.x s$ indicate that the time when the $i^{th}$ $\starl{doReachAvoid}(x[i],U)$ was executed by the robot.
As expected, the AR Drone is slow to get to $A$ as it has to first take-off, but then it reaches $B$ and $C$ much sooner.

\begin{figure}[h!]
	\begin{center}
		\includegraphics[scale=0.15]{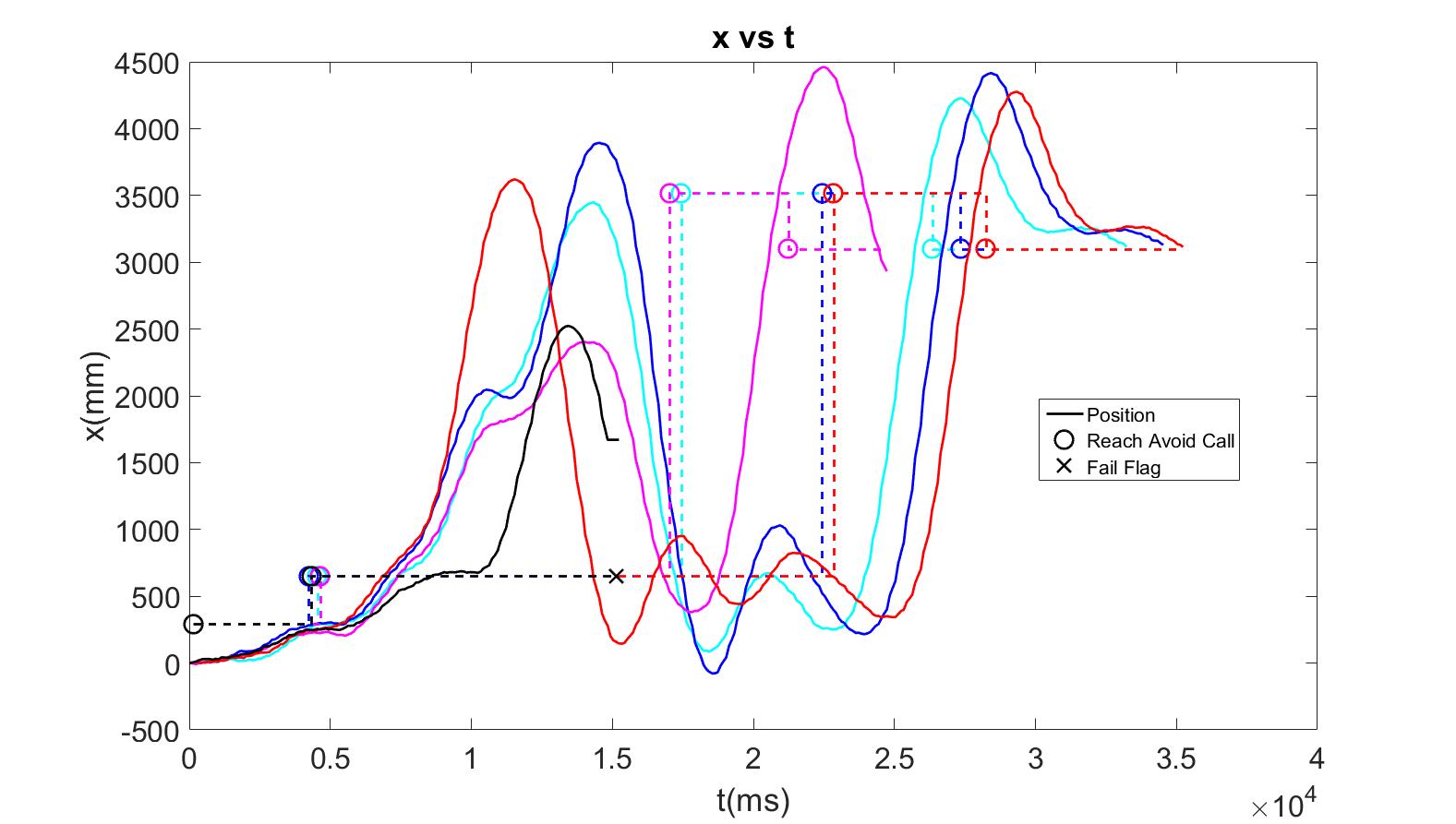}
			\includegraphics[scale=0.15]{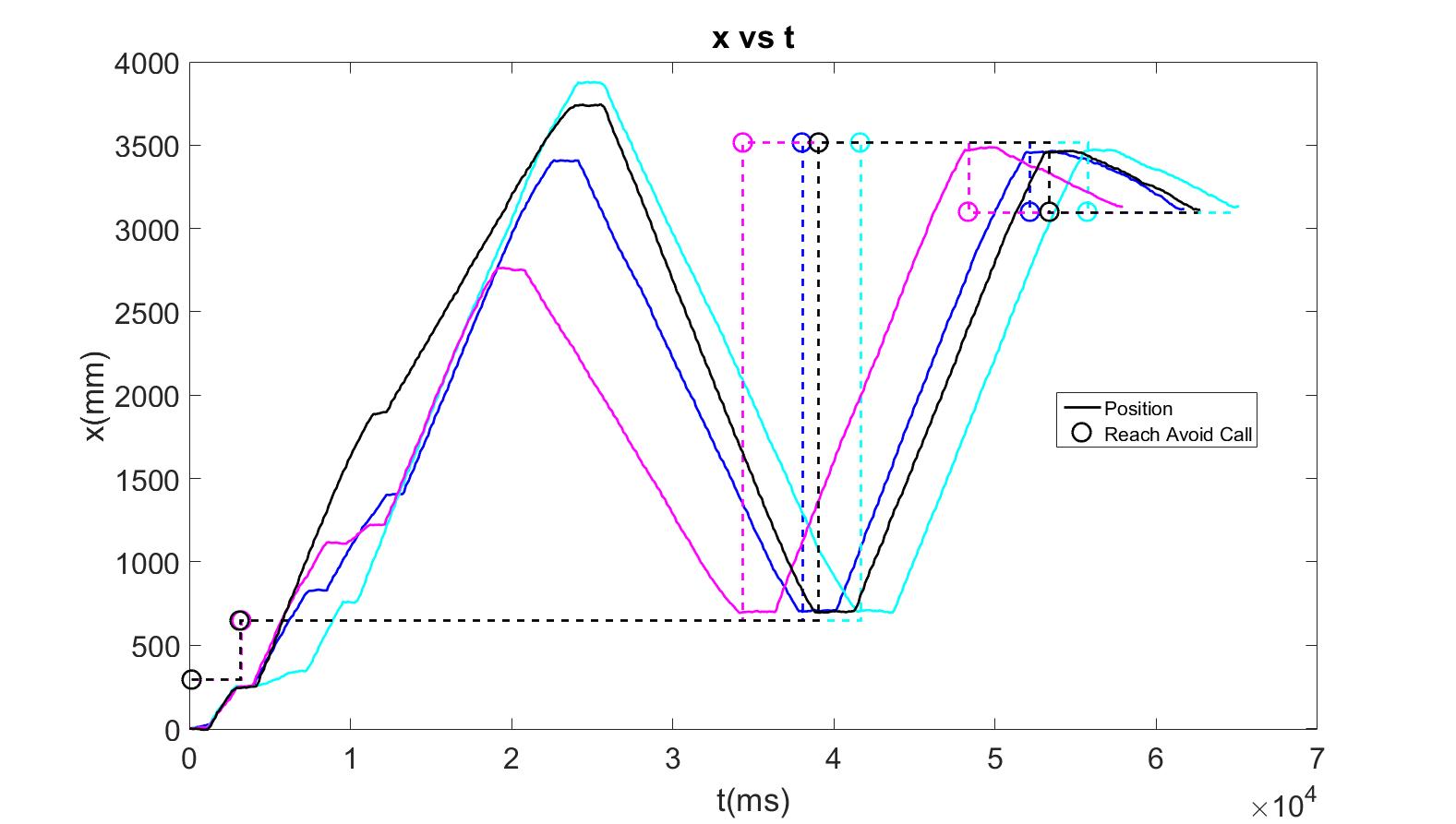}		
	\end{center}
	\caption{\scriptsize Timed traces of quadcopters (top) and iRobot (bottom). The x-coordinate (solid lines),  calls to doReachAvoid (circles), and the target x-coordinates (dashed lines). One of the quadcopter executions end with the $\mathit{failed}$ flag being set. }
	\label{fig:quadcopter}
\end{figure}

Figure~\ref{fig:quadcopter}  shows the time versus x-position plots for four different runs of the same program on two different platforms. 
There are several sources of uncertainty that make every simulation trace different, even on the same platform. For example, the RRT algorithm used by the path-planner is randomized, there is uncertainty in the scheduling of different threads, the underlying simulation models for the dynamics involves random effects such as wind, sensor noise, etc. 
Across different platforms the timing parameter $d_t$ and the quantization distance $q_d$ are different, and of course, the dynamics is also different. 

For each of these traces, once again, the hollow circles show the time of the $\starl{doReachAvoid}(x[i], U)$ call and the dashed line to its right show the $x$-coordinate of $x[i]$, that is, the next $\mathit{targetPos}$. 
Once the robot reaches the $q_d$-ball around this target position for both $x$ and the $y$ coordinate (latter is not shown in the Figures), then the $\mathit{done}$ flag is set and soon afterwards (depending on the execution time of the program thread) a new $\starl{doReachAvoid}(x[i], U)$ is called.

Observe that the quadcopter overshoots the $\mathit{targetPos}$ sometimes ($B$ in Figure~\ref{fig:both}). The black execution of the quadcopter ends with the $\mathit{fail}$ flag being set as the quadcopter drifts into  $\mathit{unsafePos}$ because of wind (Figure~\ref{fig:quadcopter}).


%

If there are moving objects blocking progress or if the $\mathit{targetPos}$ is too close to the  $\mathit{unsafePos}$, the path planner may  be unable to find a safe path in which case it sets the $\mathit{active}$ flag to false to inform the application program.


\subsection{Discussion and other applications}
\label{sec:apps-experiments}

\begin{figure}
	\begin{center}
		\scriptsize
\begin{lstlisting}[language=xyzNums]
loc: enum{init, pick, wait} = init; 
target[] : ItemPositon
currentIndex: int;
sharedmw sharedIndex: int = 0; 
 
initialize()
	pre loc == init
	eff target = $\starl{getInput}()$; currentIndex = 0; loc = pick; $\label{ln:initL}$ 

pick()
	pre sharedIndex $<$ targets.size()  $\&\&$ 
	(loc == pick || $! \starl{active}$ || $\starl{failed}$ || $\mathit{currentIndex} != \mathit{sharedIndex}$ ) $\label{ln:pick}$ 
	eff currentIndex = sharedIndex;
		$\starl{doReachAvoid}$(target[currentIndex],$\starl{getObstacles}$()); $\label{ln:move}$ 
		loc = wait;

wait()
	pre loc == wait $\&\&$ $\starl{done}$		 
	eff sharedIndex = currentIndex + 1; loc = pick; $\label{ln:reached}$ 
			
\end{lstlisting}
		\caption{\scriptsize Distributed way-point tracking application.}
		\label{fig:race}
	\end{center}
\end{figure}

While a detailed user study has been beyond the scope of the current work, we believe that the StarL language constructs, primitives, and now the run time system supporting portability,  dramatically improves programmer productivity and experience for developing distributed mobile robotic applications.  

To illustrate this, we discuss some other StarL applications. Figure~\ref{fig:race} shows the code for a distributed way-point tracking application in which  the participating robots need to collectively cover a stream of way-points. The shared variable $sharedIndex$ is used by the group to track the current way-point of interest. Each robot reads the $sharedIndex$ and stores it in the local variable $currentIndex$. Then it calls the $\starl{doReachAvoid}$ function to race to the target while avoiding obstacles. If and when the $\starl{done}$ flag becomes true, $sharedIndex$
is incremented which indicates to all the other robots that its time to move to the next way-point.
If the shared variable is changed, the program again calls  $\starl{doReachAvoid}$ function to move to the  next target. 

\begin{figure}
	\begin{center}
		\includegraphics[scale=0.2]{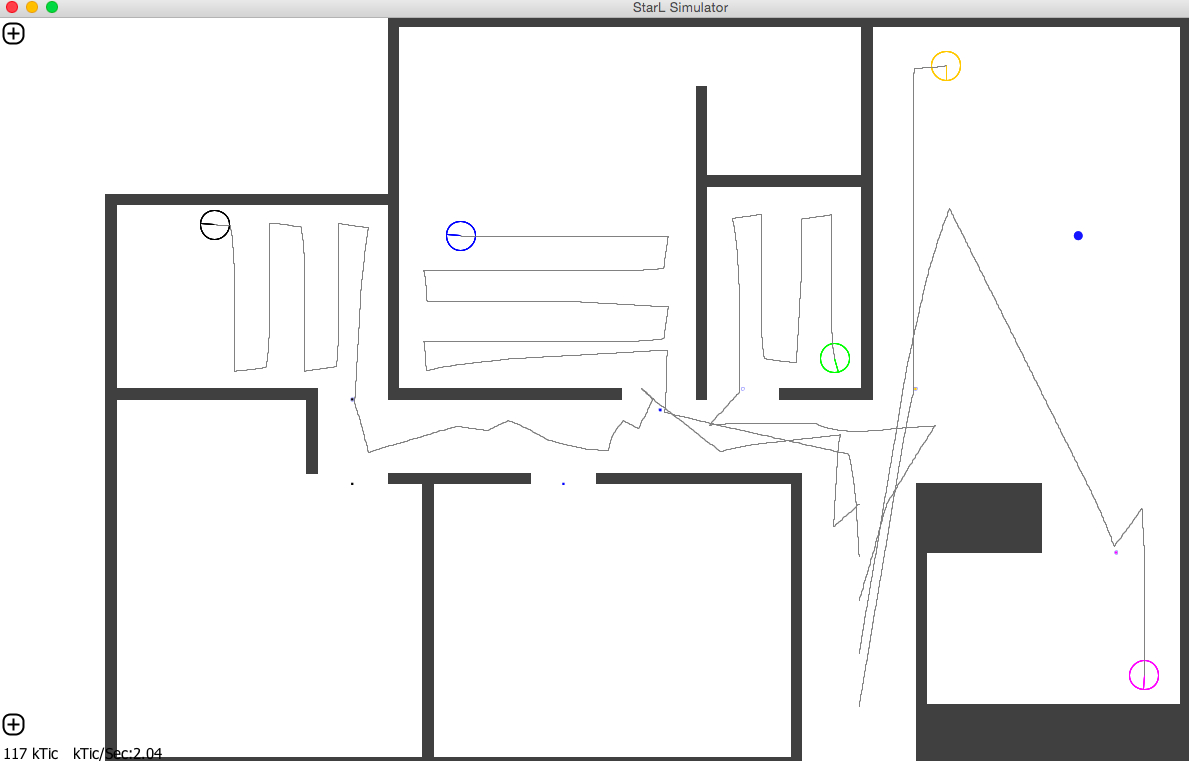}
	\end{center}
	\caption{Screen shot of Distributed Search Simulation}
	\label{fig:hetergeneos}
\end{figure}

In addition to reach-avoid and the DSM, StarL also provides primitives that are commonly used idioms in distributed systems such as  leader election, mutual exclusion, and set consensus. 
%
Programs for complex applications become simple and modular using these primitives. A number of portable demo applications for distributed robots have been developed~\cite{DBLP:conf/lctrts/LinM15}. For example, our collaborative search builds up on the distributed way-point tracking applications: The input is a map with a number of rooms to be searched. At the beginning, a leader is elected; the leader assigns the rooms (way-points) to all the robots using DSM. When each robot receives the assignment from the leader, it uses $\starl{doReachAvoid}$ to move to the entrance of the room and then search the room. Finally, it updates the  DSM to relay the status of the search to the others. A screen shot of the StarL simulator running this application is shown in Figure~\ref{fig:hetergeneos}. 

\section{Conclusions}
\label{sec:conc}
In this paper, we introduce the new StarL programming framework which supports porting of high-level application programs across different platforms.
We showed utility of the language constructs for control and coordination, namely doReachAvoid and distributed shared memory, through several example applications.
In order to discuss portability of StarL programs, we first propose a semantics of these programs. 
It is meaningless to discuss semantics of a robotic program without specific reference to timing and spatial positions and orientations. At the same time, if the semantics is too closely attached to the specific dynamics of a platform, the purpose of portability is defeated.  
In this paper, we reach a compromise and propose the semantics of StarL applications in terms of key program variables that are platform independent, and certain platform dependent parameters related to sensing precision and timing. 
This semantics then enables us to state what it means for an application program to be ported to a new platform with new parameters but identical program interfaces. 
Based on this semantics,  we present the design of the StarL runtime system that enables us to automatically port application programs. Specifically, given the dynamical model of a target platform and a low-level controller interface, the StarL compiler automatically generates executable code for the platform that meets the semantics. The key element of the runtime system is a robot controller that interfaces with the doReachAvoid construct. 
Through detailed simulations of StarL application programs on two popular platforms, we have established that the runtime system can meet the proposed semantics.

\section{Acknowledgments}

We are thankful to individuals who have contributed to the development and evaluation of the StarL platform. 
Ritwika Ghosh developed the compiler for the high-level language. 
Zhenqi Huang and Kenji Fukuda developed the way-point tracker for the AR Drone2 platform. 

\bibliographystyle{plain}

\bibliography{sayan1}

\end{document}